\documentclass{article}
\pdfoutput=1
\usepackage{amsmath,graphicx, bm}
\usepackage{pdfpages}
\usepackage[preprint]{spconf}
\usepackage{spconf}
\DeclareMathOperator*{\argmax}{arg\,max}

\usepackage{color}
\usepackage{hyperref}

\usepackage{booktabs}
\usepackage{amsfonts}
\usepackage{url}
\usepackage{xcolor,colortbl}
\definecolor{Gray}{gray}{0.85}

\title{HIERARCHICAL MULTITASK LEARNING WITH CTC}

\name{Ramon Sanabria and Florian Metze}
\address{Carnegie Mellon University\\
	Language Technologies Institute, School of Computer Science\\
	Pittsburgh, PA; U.S.A.\\
    \texttt{\{ramons{\textbar}fmetze\}@cs.cmu.edu}
    }
\begin{document}
\ninept

\newcolumntype{L}[1]{>{\raggedright\arraybackslash}p{#1}}
\newcolumntype{C}[1]{>{\centering\arraybackslash}p{#1}}
\newcolumntype{R}[1]{>{\raggedleft\arraybackslash}p{#1}}

\copyrightnotice{\copyright\ IEEE 2018}
                 \toappear{Published in the 2018 IEEE Workshop on Spoken Language Technology (SLT 2018), Athens, Greece}
%
\maketitle
\begin{abstract}



In Automatic Speech Recognition, it is still challenging to learn useful intermediate representations when using high-level (or abstract) target units such as words. For that reason, when only a few hundreds of hours of training data are available, character or phoneme-based systems tend to outperform word-based systems. In this paper, we show how Hierarchical Multitask Learning can encourage the formation of useful intermediate representations. We achieve this by performing Connectionist Temporal Classification at different levels of the network with targets of different granularity. Our model thus performs predictions in multiple scales for the same input. On the standard 300h Switchboard training setup, our hierarchical multitask architecture demonstrates improvements over singletask architectures with the same number of parameters. Our model obtains 14.0\% Word Error Rate on the Switchboard subset of the Eval2000 test set without any decoder or language model, outperforming the current state-of-the-art on non-autoregressive Acoustic-to-Word models.

\end{abstract}
\section{Introduction}
\label{sec:intro}




Automatic Speech Recognition (ASR) systems, traditionally, were composed of three elements: a phoneme-based Acoustic Model (AM), a word-based Language Model (LM), and a pronunciation lexicon that maps a sequence of phonemes to words~\cite{jurafsky2000speech}. This setup was predominating for systems based on Gaussian Mixture Models (GMMs) and Hidden Markov Models (HMMs)~\cite{rabiner1986introduction}. More recently, systems that used a single loss-function based on the predicted sequence (\textit{i.e.},~end-to-end models) showed similar performance by using characters as target units. More specifically, these approaches use either  a Sequence-to-Sequence (S2S) architecture~\cite{chiu2017state} or a Connectionist Temporal Classification (CTC) loss-function~\cite{miao2015eesen}. By not requiring a pronunciation lexicon anymore, they simplify the ASR pipeline and obtain competitive results.

\begin{figure}[t!]
 \centering
 \includegraphics[width=\linewidth]{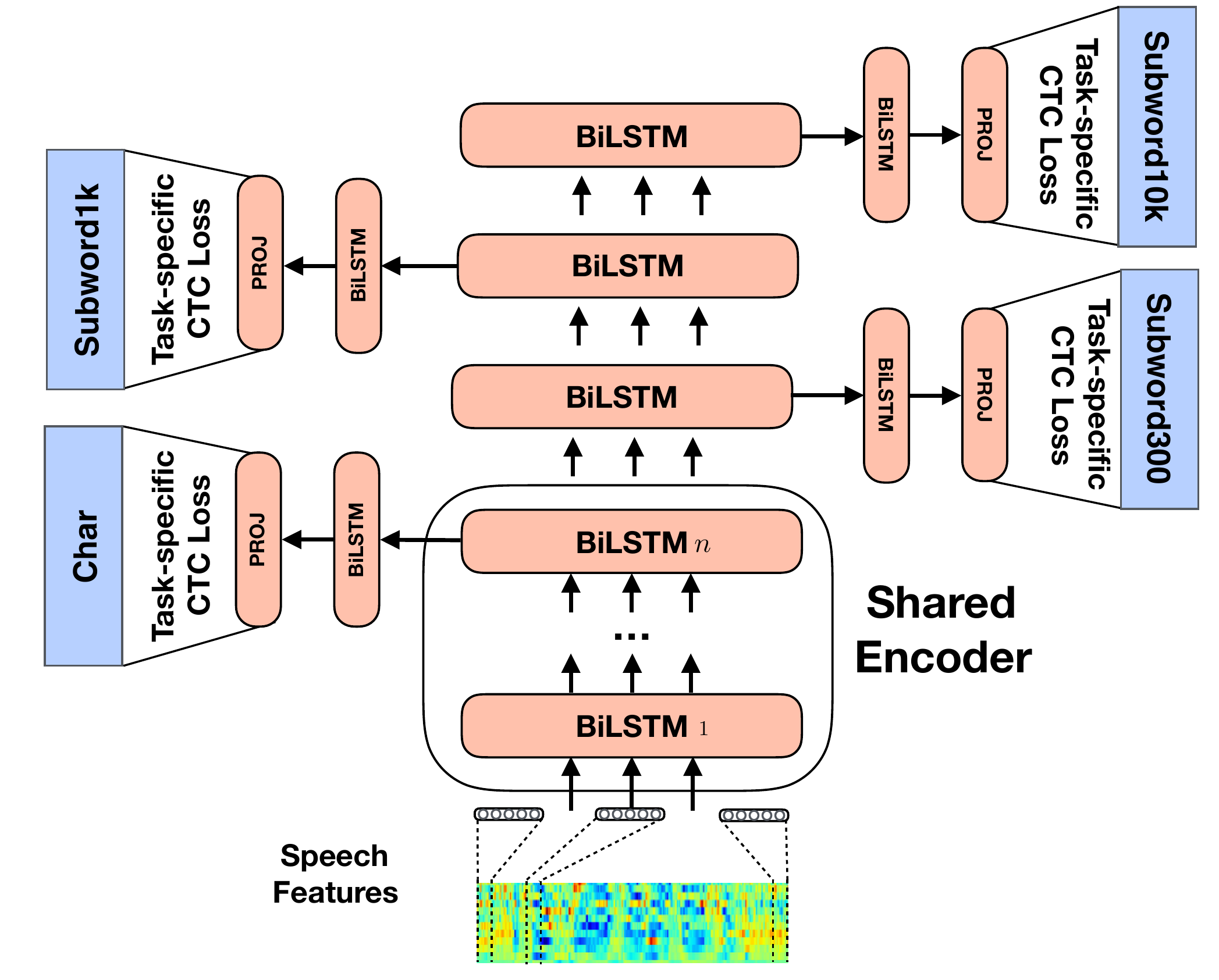}
  \caption{Our Hierarchical Multitask Learning (HMTL) Model learns to recognize word-level units with a cascade of fine-to-coarse granularity CTC auxiliary objectives, progressing from recognizing character-level units to almost word-level units at the Subword10K CTC loss.}\label{label:hmtl}
\end{figure}

To simplify even more the ASR pipeline, Soltau \textit{et al.} suggest using words instead of characters as target units~\cite{soltau2016neural}. Even though this technique forces a closed vocabulary setting, experiments show competitive results on a large training corpus. Audhkhasi~\textit{et al.} test the same approach on Switchboard, a smaller dataset, and show that the model can only predict reliably words that frequently appear in the training set~\cite{audhkhasi2017direct}. Concerning this issue, multiple approaches use intermediate level target units (between characters and words) such as subwords~\cite{sennrich2015neural} that provide a more balanced solution. 

In this work, we introduce a new Hierarchical Multitask Learning architecture (HMTL) (see Fig.~\ref{label:hmtl}) that maintains the modularity of traditional models by using a Multitask Learning (MTL)~\cite{caruana1997multitask} strategy without giving up the simplicity of end-to-end approaches. Our method exploits the inherent compositionality of different unit representations. By providing auxiliary loss-functions, we hypothesize that the network is forced to learn useful intermediate representations. The experimental observation that each module in a deep neural network architecture learns (hierarchically) specific representations of the input signal inspired the design of this model.

Our experiments show that this technique outperforms comparable singletask learning architectures (STL) and traditional MTL techniques. We also establish the current state-of-the-art on non-autoregressive acoustic-to-word models (A2W)~\cite{audhkhasi2017building} on the 300 hours Switchboard subset without exhaustive hyper-parameter optimization. 


\section{Related Work}
\label{sec:format}


During the last decade of development on end-to-end speech recognition systems~\cite{bahdanau2016end,graves2006connectionist}, the ASR community started paying more attention to different types of target units. In this direction, Soltau~\textit{et al.} propose a CTC model that uses words as target units without any LM~\cite{soltau2016neural}. Later, Audhkhasi~\textit{et al.} port this idea to smaller and public datasets achieving competitive results~\cite{audhkhasi2017building,audhkhasi2017direct}. 

The high number of dataset-specific hyper-parameter tuning needed (\textit{e.g.}, number of layers, optimizers, learning rate scheduler) for A2W models makes difficult the incorporation of target units as an extra hyper-parameter.
It is still an open question which is the optimal unit length or the number of targets for end-to-end models. Small target units give more flexibility, and they are complicated to decode. On the other hand, big units are less flexible and usually easier and faster to decode.
In~\cite{zenkel2017subword} we observe that each type of target unit contributes differently to the final output. As a possible solution, Chan \textit{et al.} and Liu \textit{et al.} propose to learn the best possible decomposition of units~\cite{chan2016latent,liu2017gram}. 


In a different direction, some approaches propose to combine the hypothesis of various models that use different sets of target units. For instance, Hori \textit{et al.} suggest to combine the predictions of a word-based LM with the ones from a character-based during decoding~\cite{hori2017multi}. The results obtained show improvements over character-only decoding. Another example is~\cite{zenkel2017subword}, where we show that by combining hypothesis~\cite{fiscus1997post} from various models that use different targets units, the final output recovers some of the mistakes done by the individual models.


Another potential solution to take into consideration different unit representations is to use MTL techniques. Recently, the ASR community started showing interest in MTL approaches. One example of this recent interest is~\cite{kim2017joint}. In this work, Kim \textit{et al.} combines a S2S with a CTC-loss by using MTL techniques. In this case, the CTC-loss was only used as an auxiliary loss-function to enforce a monotonic alignment between the speech signal and the output label sequence. Another ASR task where MTL is applied is in multilingual speech recognition. In this direction, we propose~\cite{dalmia2018sequence}. This work shows that in a common MTL setting, the network can learn useful intermediate representations for multiple languages. By using various CTC-losses, our architecture was able to learn useful intermediate representations common among all inputs.

MTL has also been used hierarchically to provide low-level supervision to deep neural networks. In this direction, S{\o}gaard and Goldberg propose an architecture that learns more fundamental Natural Language Processing tasks to guide the internal representation of a neural network~\cite{sogaard2016deep}. Focused on ASR, Fernandez \textit{et al.} use mono phone prediction as an auxiliary task to improve digit recognition using CTC~\cite{fernandez2007sequence} . In this case, the auxiliary loss helps the network to learn an intermediate phonetic representation. More recently, Toshniwal \textit{et al.} and Krishna \textit{et al.} port this idea to a more realistic dataset and use characters and workpieces as higher level units respectively~\cite{toshniwal2017multitask, krishna2018hierarchical}. There are three fundamental differences between~\cite{fernandez2007sequence, toshniwal2017multitask, krishna2018hierarchical} and our work. First, we analyze and show improvements in different multitask learning architectures. Second, our method takes advantage of the inherent compositionality of the units rather than the acoustic characteristics of the signal. The fact of not considering acoustics, at the same time, allows us to not rely on phoneme annotations. Working with phonemes requires a dictionary which maps a word to its phonetic transcriptions. Such transcriptions are expensive and prone to error. Third, our method can use a higher number of auxiliary tasks.

\section{Unit Selection}
\label{sec:unit}

Different methods have been proposed to create intermediate target representations between character and words. For instance, Liu~\textit{et al.} use bigram and trigram units constructed from characters by a CTC model~\cite{liu2017gram}. Another example is~\cite{sennrich2015neural}. In this work, Sennrich~\textit{et al.} use the Byte-Pair Encoding (BPE) algorithm~\cite{gage1994new} to overcome the fixed length restriction applied in~\cite{liu2017gram}. By using BPE, they create units according to the frequency of repetition fixing the number of targets. This approach has been extensively used by the Machine Translation community (MT) and for LM~\cite{mikolov2012subword}. The inherent compositionality of BPE sets (\textit{i.e.},~units with fewer operations form units with more BPE operations) makes this technique an appropriate candidate for HMTL, the model proposed in this work.

\begin{figure}[t!]
 \centering
 \includegraphics[width=\linewidth]{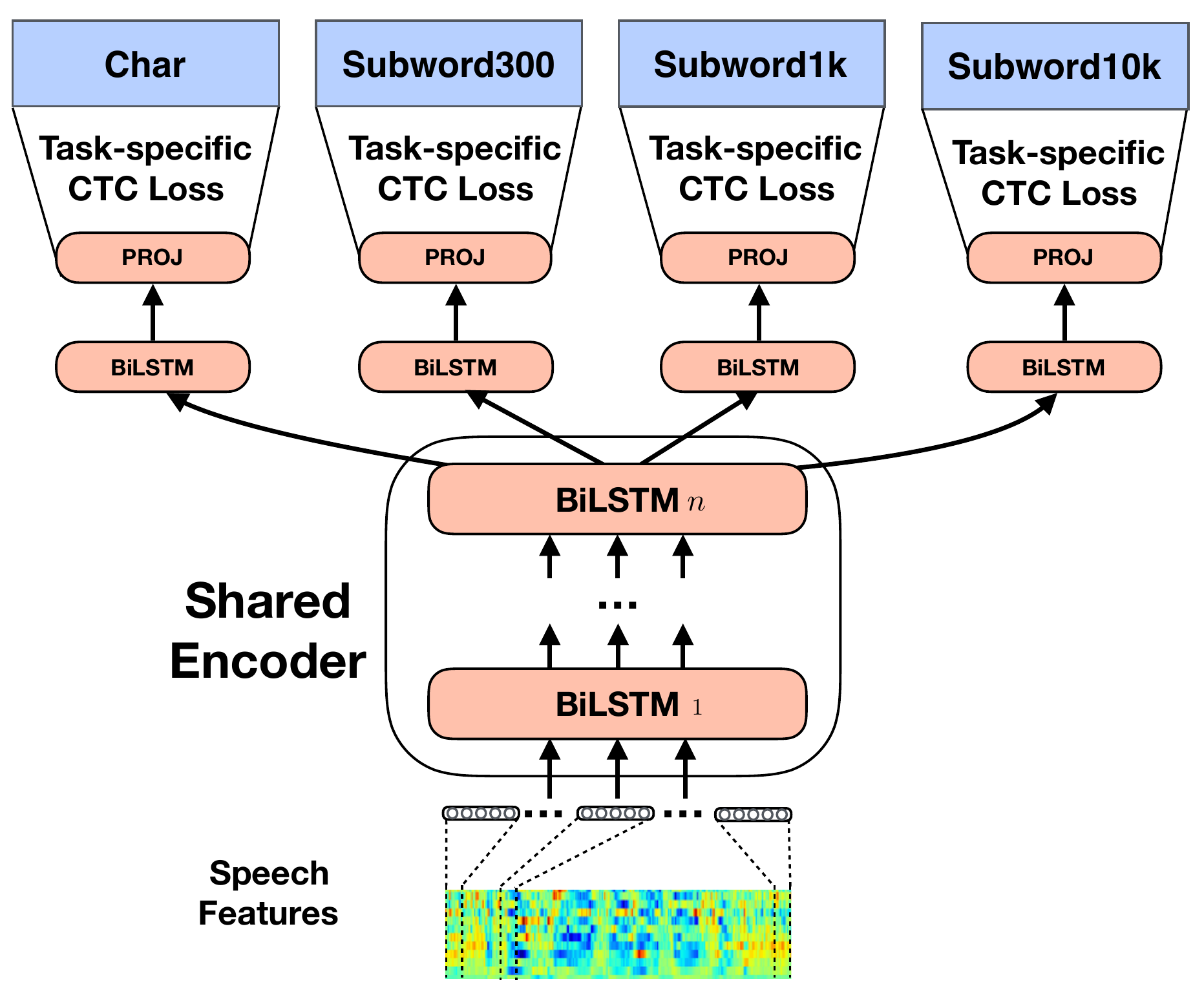}
   \caption{The more common Block Multitask Learning (BMTL) model learns intermediate representation using its shared encoder. The task-specific modules will use this intermediate representation to discriminate each unit.}\label{label:bmtl}
\end{figure}

\subsection{Byte-Pair Encoding Algorithm}

Gage propose the BPE algorithm as a compression algorithm~\cite{gage1994new}. This technique iteratively replaces pairs of most frequently co-occurrent bytes creating a new byte. A table that maps the correspondence of the new unit and its composition is needed to recover the original form of the data.

We apply this algorithm to natural languages by treating each target unit (\textit{e.g.},~a character) as a byte. More specifically, in our case, our initial unit set (\textit{i.e.},~our first set of bytes) are characters~(see Section~\ref{sec:char}). Each BPE step/operation performs a unit merging. For that reason, the number of BPE operations roughly defines the number of units of the target set. For instance, if two targets in our vocabulary are `HELL' and `O' and they appear together frequently, by applying a BPE operation we can merge them in a new target unit: `HELLO'. This operation can be repeated an arbitrary number of times until converging to a target set that contains all words of the vocabulary.

\subsection{Subword Units}
~\label{sed:subword}
Subwords are the units created from characters by using BPE operations. More concretely, we use~\cite{sennrich2015neural} to create our Subword units. To generate different units we start with a character set. We also use a unique character (in our case `@') to determine the position of the Subword inside a word. We do not use a dedicated space character because we define word boundaries by the absence of the token `@' within a unit. 

We create three different sets of subword units: Subword300 (s300), Subword1k (s1k) and Subword10k (s10k). The number after `Subword' defines the number of BPE operations used to create that set, that is roughly the number of units that the set has. More precisely, taking the \emph{blank} symbol for CTC computation (see Section~\ref{sec:CTC}) into consideration, each set contains 388, 1,088 and 10,088 target units respectively. Table~\ref{tab:split} shows an utterance from Switchboard dataset (see Section~\ref{sec:results}) decomposed using different BPE operations. Note that as the number of BPE operations increases, the amount of units that represents whole words also increases. For that reason, we can consider s10k an approximation of a word target set.


\begin{table}[t!]
  \centering
  \begin{tabular}{ c  p{5.6cm} }
  Method & Utterance \\
  \hline
  Original & you know it's no not even cold weather \\
  Character & y o u k n o w i t ' s n o n o t e v e n c o l d w e a t h e r \\
  Subword 300 & you know it's no not even co@ ld w@ ea@ ther \\
    Subword 1k & you know it's no not even co@ ld wea@ ther  \\
  Subword 10k & you know it's no not even cold weather 
\end{tabular}
  \caption{Utterance sw02054-B\_032568-033031 from the Switchboard dataset decomposed into characters, 300, 1,000 and 10,0000 BPE operations. The character `@' denotes that the unit is in the middle of a word.}\label{tab:split}
\end{table}

\subsection{Characters}
\label{sec:char}

The character set is the base set from where we construct the Subword units. It contains alpha-numerical characters, punctuation and the \emph{blank} symbol: 48 symbols in total.

Because we can use the position symbol `@' to find word boundaries when using subword units (\textit{i.e.},~s300, s1k, s10k) and to stress the compositionality of our model, we do not include a `space' representation in our character set. Therefore, the character representation of our model predicts strings of characters without spaces as we can see in Table~\ref{tab:split}.

\section{Multitask Learning Architecture}
\label{sec:mtarch}

In this section, we introduce two different architectures that accept auxiliary loss-functions. First, we present Block Multitask Learning (BMTL) (see Fig.~\ref{label:bmtl}), the conventional approach for multitask learning with CTC inspired by our previous work~\cite{dalmia2018sequence}. Second, we introduce our new HMTL architecture (see Fig.~\ref{label:hmtl}) that is inspired by~\cite{toshniwal2017multitask,fernandez2007sequence,krishna2018hierarchical} and takes advantage of the compositionality of subwords. Although both models use CTC as auxiliary loss-functions, we can port them to S2S models.

\subsection{Singletask Learning Baseline}

The STL baseline architecture is similar to the approach that we propose in ~\cite{zenkel2017comparison,zenkel2017subword}. The acoustic encoder reads a sequence of acoustic features vectors $\bm{X} = (\bm{x}_0 \cdots \bm{x}_T$), where $\bm{X} \in \mathbb{R}^{F \times T}$, $F \in \mathbb{N}^{+}$ is the dimensionality of the feature vector and $T \in \mathbb{N}^{+}$ the number of frames of the audio sequence. Then, a Bidirectional Long-short Term Memory (BiLSTM)~\cite{hochreiter1997long} takes $\bm{X}$ as input and returns a matrix $\bm{H} \in \mathbb{R}^{(2 \times H) \times T}$ where $H \in \mathbb{N}^{+}$ is the dimensionality of one direction of a hidden state of the BiLSTM. Then, each hidden state  is projected by a function $\mathcal{F}: \mathbb{R}^{(2 \times H)\times T} \rightarrow \mathbb{R}^{|L'|\times T}$, where $L' = L \cup \emptyset$ is the extended character set, $L$ is our original character set, and $\emptyset$ is the \emph{blank} symbol. Note that $|L'| = |L|+1$. 

Finally, we apply a softmax function over each row $\mathcal{S}: \mathbb{R}^{|L'|\times T} \rightarrow \{ \bm{H} \in \mathbb{R}^{|L'|\times T}\}$ where $0<\bm{H}_{l',t} <1$, creating a probability distribution over the extended unit target set $L'$ for each frame. More formally, $\sum_{l'=0}^{|L'|}\bm{H}_{l',t} = 1$.  Then, CTC uses this matrix and dynamic programming techniques to compute the probability of generating all possible correct sequences.

\subsubsection{CTC}
~\label{sec:CTC}

Let $\bm{z} = (z_1, \cdots ,z_U) \in \mathbb{N}^{+U}$ be a ground truth sequence where $U\in \mathbb{N}^{+}$, $U \leq T$ and $z_u < |L| $. In this paper, $\bm{z}$ is the transcription of an input sequence. Let us define a many to one mapping function  $\mathcal{B}: \mathbb{N}^{+T} \rightarrow \mathbb{N}^{+U}$, where $T \in \mathbb{N}^{+}$ is the input length of the feature matrix $\bm{X}$. $\mathcal{B}$ will therefore map a sequence $\bm{p} = (p_1, \cdots, p_T) \in \mathbb{N}^{+T}$, where $ p_t < |L'|$, to the ground truth sequence $\bm{z}$.  

This mapping $\mathcal{B}$ is also referred as the squash function, as it removes all \emph{blank} symbols of the path and squashes multiple repeated characters into a single one (\textit{e.g.},~$\mathcal{B}(AA \emptyset AAA BB) = AAB$). Note that we do not squash characters that are separated by the \emph{blank} symbol because this allows to predict repeated characters. Then, let us define the probability of the sequence $\bm{p}$ as
\begin{equation}
P(\bm{p}|\bm{X}) = \prod_{t=1}^T P(p_t | \bm{x})
\end{equation}
where $P(p_t | \bm{x})$ is the probability of observing the label $p_t$ at time~$t$. To calculate the probability of an output sequence $\bm{z}$ we sum over all possible sequences:
\begin{equation}
P(\bm{z}|\bm{X}) = \sum_{\bm{p}\in \textit{B}^{-1}(\bm{z})}P(\bm{p}|\bm{X})
\end{equation}
To sum over all sequences, we use a technique inspired by the traditional dynamic programming method used in HMMs: the forward-backward algorithm~\cite{rabiner1986introduction}. We additionally force the appearance of \emph{blank} symbols in the sequence $\bm{p}$ by augmenting it with \emph{blank} symbols between each of the labels of $\bm{z}$ as well as at the beginning and the end of the sequence. 

Given a sequence of speech features $\bm{X}$, we can now compute the probability that the network correctly predicts the ground truth sequence $\bm{z}$. In the following subsections, we present decoding strategies that process the output of the system trained with CTC to create a linguistically reasonable transcription.

\subsubsection{Greedy Search}

To create a transcription without adding any linguistic information we can greedily search  the best sequence $\bm{p} \in \mathbb{N}^{+T}$:
\begin{equation}
\argmax_{\bm{p}} \prod_{t=1}^T P(p_t|\bm{X})
\end{equation}
The mapping of the path to a transcription $\bm{z}$ is straight forward.  We can map both sequences by applying the squash function: $\bm{z} = \mathcal{B}(\bm{p})$.

\subsubsection{Shallow Fusion}
\label{subsec:shallow}

We can combine a unit-specific LM with the probability distribution given by a CTC-trained model. Let us assume that the alphabet of the character based LM is equal to $L$. We want to find a transcription that has a high probability based on the acoustics from the CTC model as well as the LM. Since we have to sum over all possible sequences $\bm{p}$ for a transcription $\bm{z}$, our goal is to solve the following equation:
\begin{equation}
\argmax_{\bm{z}} \sum_{\mathcal{B}^{-1}(\bm{z})} \prod_{t=1}^{T}
P_{AM}(p_t|\bm{X})\times P_{LM}(p_t|\mathcal{B}(\bm{p}_{1:t-1}))
\end{equation}
Where $P_{AM}$ is the probability of the AM and $P_{LM}$ is the probability of the LM. Note that we can not estimate a useful probability for the \emph{blank} label $\emptyset$ with the LM, so we set $P'_{LM}(\emptyset | \mathcal{B}(\bm{p}_{1:t-1})) = 1 $. To not favor a sequence of \emph{blank} symbols, we apply an insertion bonus $b \in \mathbb{R}$ for every $p_t \neq \emptyset$ as:
\begin{equation}
P_{LM}(k|\mathcal{B}(\bm{p}_{1:t-1}))= 
\begin{cases}
    P_{LM}(k|\mathcal{B}(\bm{p}_{1:t-1})) \cdot b,& \text{if } k \neq \emptyset\\
    1,              & \text{if } k = \emptyset
\end{cases}
\label{eq:beam}
\end{equation}
Where $P_{LM}(k|\mathcal{B}(\bm{p}_{1:t-1}))$ is provided by the unit-specific LM. Because it is unfeasible to calculate an exact solution to equation~(\ref{eq:beam}) by brute force, we apply beam search similar to \cite{hwang2016character}.

Shallow fusion is compatible with any target unit. More concretely, we apply this approach to subword, character or word unit systems. Note that this technique is independent of the architecture of the AM. For instance, in this work, we apply shallow fusion to STL or MTL architecture.

\subsection{Block Multi Task Learning}
\label{sec:bmtl}

BMTL (see Fig.~\ref{label:bmtl}) is a general architecture for multitask learning. This approach shares a $n$ BiLSTM layer encoder until before the unit-specific modules. After that, a unit-specific module process the hidden representations of the shared encoder. We compose this module with one BiLSTM layer and a projection layer that maps the hidden states of the BiLSTM layer to the dimensionality of each subword/character unit set plus one. 

More formally, BMTL can be written as,

\begin{equation}
\begin{split}
\bm{e}_0 &= Shared\_Encoder(\bm{X})\\
\bm{L}_{char} &= softmax(\bm{W}_{char} BiLSTM_{char}(\bm{e}_0))\\
\bm{L}_{s300} &= softmax(\bm{W}_{s300} BiLSTM_{s300}(\bm{e}_0))\\
\bm{L}_{s1k} &= softmax(\bm{W}_{s1k} BiLSTM_{s1k}(\bm{e}_0))\\
\bm{L}_{s10k}&= softmax(\bm{W}_{s10k} BiLSTM_{s10k}(\bm{e}_0))\\
\end{split}
\end{equation}

Where $Shared\_Encoder$ function is a $n$ layer BiLSTM that is shared across all unit-specific modules, $\bm{e}_0$ is the hidden representation generated by the shared encoder, $BiLSTM_{n}$ with $n\in \{char,s300,s1k,s10k\}$, is a one layer unit-specific BiLSTM, $\mathbf{W} \in \mathbb{R}^{|L_n'|\times ( 2\times H)}$ represents a unit-specific linear transformation (feedforward network) that maps the bidirectional hidden dimensionality of the unit-specific BiLSTM  to the actual target unit dimension plus one, $\bm{X} \in \mathbb{R}^{F \times T}$ is the input sequence feature vectors and $\bm{L}_{n} \in \mathbb{R}^{|L_n'| \times T}$ is the probability matrix of each frame over each target unit set $n$ plus one. Note that the output matrix $\bm{L}_{n}$ is then processed either for CTC during training or for a decoder during testing.

Because the model needs to generate different output representations of the same input, we hypothesize that the shared encoder will learn a more general and flexible encoding of the audio features. Finally, we assume that the subword/character-specific modules will learn how to discriminate each unit separately using the representation generated by the shared encoder.

\subsection{Hierarchical Multi Task Learning}

HMTL (see Fig.~\ref{label:hmtl}), similar to BMTL, shares an encoder composed of $n$ BiLSTM layers. The fundamental difference between BMTL and HMTL is that HMTL will learn intermediate representations in a fine-to-coarse fashion after each layer. More concretely, following the encoder, a cascade of BiLSTM layers are stacked on the model. After each layer, we connect an auxiliary task-specific module. A BiLSTM and a linear projection layer form this task-specific module that maps the hidden representation of each layer to the specific number of units of each target set plus one. The projection made by the task-specific module allows us to compute CTC in different parts of the network.

More formally, we can define the main structure of HMTL as

\begin{equation}
\begin{split}
\bm{e}_0 &= Shared\_Encoder(\bm{X})\\
\bm{e}_1 &=  BiLSTM_{1}(\bm{e}_0)\\
\bm{e}_2 &=  BiLSTM_{2}(\bm{e}_1)\\
\bm{e}_3 &=  BiLSTM_{3}(\bm{e}_2)\\
\end{split}
\end{equation}

Then, we will obtain a probability distribution by using unit-specific module as

\begin{equation}
\begin{split}
\bm{L}_{char} &= softmax(\bm{W}_{char} BiLSTM_{char}(\bm{e}_0))\\
\bm{L}_{s300} &= softmax(\bm{W}_{s300} BiLSTM_{s300}(\bm{e}_1))\\
\bm{L}_{s1k} &= softmax(\bm{W}_{s1k} BiLSTM_{s1k}(\bm{e}_2))\\
\bm{L}_{s10k} &= softmax(\bm{W}_{s10k} BiLSTM_{s10k}(\bm{e}_3))\\
\end{split}
\end{equation}

Note that we use the same name convention as in Section~\ref{sec:bmtl}.



We hypothesize that this training strategy will help the model to learn how to exploit the inherent compositionality of each subwords unit.

\begin{table*}[t]
  \centering
  \begin{tabular}{ c|| c c || c c || c c || c c }
    \hline

    \% WER & STL & HMTL & STL & BMTL & STL (LM) & HMTL (LM) & STL (LM) &   BMTL (lm)\\
    \hline
        \hline

    \rowcolor{Gray}

    Subword300 &  20.6 / 33.5 & \textbf{17.6} / \textbf{30.4} &  20.6 / 33.5  & \textbf{18.7} / \textbf{31.8} & 14.2 / 26.3 & \textbf{13.1} / \textbf{24.7} & 14.2 / 26.3 & \textbf{13.5} / \textbf{25.6}\\
  
        Subword1k & 19.0 / 30.8 & \textbf{15.3} / \textbf{26.9} & 20.5 / 32.5  & \textbf{16.7} / \textbf{28.6} & 15.2 / 26.8 & \textbf{12.7} / \textbf{23.8} & 15.5 / 27.5  & \textbf{13.1} / \textbf{24.5} \\
        
       \rowcolor{Gray}
     
        Subword10k & 17.4 / 28.5 & \textbf{14.0} / \textbf{25.5} & 17.8 / 29.2   & \textbf{14.6} / \textbf{25.8} &  15.6 / 26.7& \textbf{12.5} / \textbf{23.7} & 16.2 / 27.3  & \textbf{13.1} / \textbf{24.7}\\ \hline

    \hline
  \end{tabular}
  
  \caption{ \% WER summary of the models presented in this paper on the Eval2000 test set. The AM is trained on the 300 hours subset of the Switchboard training set, while the LM is trained on the Switchboard and Fisher training sets. The results on the Switchboard subset (swbd) are on the left side of the slash and the results on the Callhome (cllhm) subset are on the right side of the slash. STL stands for singletask learning model, HMTL stands for Hierarchical Multitask Learning, BMTL stands for Block Multitask Learning and S300, S1k and S10k are the different subword target units used. All models have a two layer shared encoder. Each STL model has the same number of parameters and configuration than the model on the right. Note that, except in Subword300 unit set, HMTL and BMTL do not have the same number of parameters so they can not be compared.}~\label{taball}
\end{table*}

\section{Results}
\label{sec:results}

In this work, we use the 300 hours Switchboard subset for training the AM and the Fisher transcripts dataset for training the unit-specific LM for shallow fusion. We use the 2000 HUB5 Eval2000 (LDC2002S09) for evaluation. The corpus is composed of English telephone conversations and is divided into the Switchboard subset, which more closely resembles the training data, and the Callhome subset. We extracted 43-dimensional filter bank, and pitch features vectors with Cepstral Mean Normalization using Kaldi~\cite{povey2011kaldi}. We use the Tensorflow branch of EESEN~\cite{miao2015eesen} to develop the rest of the pipeline.

\subsection{Architecture}
\label{subsec:data}

All MTL architectures presented in this section (\textit{i.e.},~HMTL and BMTL) have a two-layer shared encoder, and each layer has 320 cells. We perform a 3-fold data augmentation by sub-sampling. We triple the number of sequences by reducing the frame rate of each sample by 3. The frames dropped during sub-sampling are concatenated to the middle frame for providing additional context. We also add a projection of 340 cells between each layer of the shared encoder. The rest of the architecture is described in Section~\ref{sec:mtarch}. Our models are decoded using greedy search or shallow fusion with a unit-specific LM composed by two unidirectional LSTM layers. The results presented in this section have not been properly hyper-parameter tuned, which may leave room for more improvement.


\subsection{Evaluation of multitask learning architectures}

In Table~\ref{taball}, we show a comparison between the STL and the MTL models presented in this paper. In this table, each pair of STL and MTL model have the same number of parameters. Note that, except in the case of Subword300, HMTL and BMTL does not have the same amount of parameters (see Section~\ref{sec:mtarch}), so they are not comparable.

In the left side of Table~\ref{taball}, we summarize all results obtained using only the AM without considering the LM during decoding. Those models are also known as A2W. We observe that in general, the addition of auxiliary tasks (while maintaining the same number of parameters) helps concerning WER. We can see that the relative improvements in HMTL (14-20\% relative WER) are higher than in BMTL (9-18\% relative WER).

The right side of Table~\ref{taball} summarizes all results obtained combining the AM and the LM probabilities by shallow fusion. We can see that shallow fusion generally helps regarding WER. The improvements of the addition of a LM, however, banish as the number of units increases. This behavior that we also observe in~\cite{zenkel2017subword}, is consistent in MTL models. 

The improvement of shallow fusion in an intermediate representation (Subword1k in STL(lm) in the next-to-last column of Table~\ref{taball}) is comparable to HMTL (Subword1k in HMTL in the first column of Table~\ref{taball}). Considering these results, we hypothesize that the addition of a higher order unit supervision on upper layers provides linguistic knowledge to the AM. 

\begin{table}
\centering
\begin{tabular}{ c c c c }
    \hline
 Model &  Type & LM & \% WER (swbd / cllhm)  \\ 
        \hline
        \hline
        \rowcolor{Gray}
  
                          Zeyer \textit{et al.}~\cite{zeyer2018improved}& S2S & NO & 13.5 / 27.4 \\ 

             Zeyer \textit{et al.}~\cite{zeyer2018improved}& S2S & YES & \textbf{11.8} / \textbf{25.7} \\                                                                         \hline
        \hline

 Ours &  CTC & NO & 14.0 / 27.1\\  
                                                                  \rowcolor{Gray}

   Ours &  CTC & YES & \textbf{12.5} / 23.7\\

     Zenkel \textit{et al.}~\cite{zenkel2017subword}& CTC & NO & 17.8 / 29.0 \\ 
                                                                 \rowcolor{Gray}

          Zenkel \textit{et al.}~\cite{zenkel2017subword}& CTC & YES & 14.7 / 26.2 \\ 

Audhkhasi \textit{et al.} ~\cite{audhkhasi2017building} & CTC & NO & 14.6 / \textbf{23.6}\\

\hline

\end{tabular}\caption{Comparison table for different approaches of A2W models. All models have been trained on 300h Switchboard and tested on the Eval2000 test set. The results on the Switchboard subset (swbd) are in the left side of the slash and the results from Callhome subset (cllhm) are in the right side of the slash.
}\label{tab:comp}
\end{table}

Finally, in Table~\ref{tab:comp}, we compare existing A2W models that, to the best of our knowledge, achieve the best results in the Eval2000 test set. We can see that HMTL outperforms previous approaches based on CTC in the Switchboard subset. Audhkhasi~\textit{et al}. \cite{audhkhasi2017building}, however, achieve a better score in the Callhome subset than our model. This improvement is most probably an indicator that HMTL is not able to generalize as good as~\cite{audhkhasi2017building} across domains. To improve the generalization of HMTL, we can apply regularization techniques such as Dropout. The approach presented by Zeyer~\textit{et al}.~\cite{zeyer2018improved}, however, outperforms other methods, with the downside that is autoregressive.

\section{Conclusions}
\label{sec:conclusion}

In this paper, we present a CTC-based Acoustic-to-Word (A2W) model with Hierarchical Multitask Learning (HMTL). This approach encourages the formation of useful intermediate representations by using targets with a different number of Byte-Pair Encoding (BPE) operations at different levels in a single network. At the lower layers, the model predicts few and general targets, while at the higher layers, the model predicts highly specific and abstract targets, such as words.

By using the Subword10k representation, after shallow fusion with a unit-specific LM, our HMTL model improves from 15.6\% WER (STL model with the same number of parameters) to 12.5\% WER. By contrast, a more traditional Multitask Learning Technique (MLT), called Block Multitask Learning (BMTL), using again Subword10k improves from 16.2\% WER (STL model with the same number of parameters) to 13.1\% WER. None of the experiments required any fine parameter-tuning.

Given that the presented model generates different-granularity outputs, we are attempting to combine the individual hypotheses~\cite{fiscus1997post} and develop a decoding scheme that jointly processes the different outputs, and integrates them directly into a word hypothesis.

\section{Acknowledgments}
\label{sec:ack}

We gratefully acknowledge the support of NVIDIA Corporation with the donation of the Titan X Pascal GPU used for this research. The authors would also like to thank the anonymous reviewers, Karen Livescu, Thomas Zenkel, Siddharth Dalmia, Jindrich Libovicky, Ozan Caglayan, Amanda Duarte, Elizabeth Salesky,  Desmond Elliott and Pranava Madhyastha for sharing their insights. 


\bibliographystyle{IEEEbib}
\bibliography{refs}

\begin{thebibliography}{10}

\bibitem{jurafsky2000speech}
Dan Jurafsky and James Martin,
\newblock {\em Speech \& language processing},
\newblock Pearson Education, 2000.

\bibitem{rabiner1986introduction}
Lawrence Rabiner and B~Juang,
\newblock ``An introduction to hidden markov models,''
\newblock {\em ASSP Magazine}, 1986.

\bibitem{chiu2017state}
Chung-Cheng Chiu, Tara~N Sainath, Yonghui Wu, Rohit Prabhavalkar, Patrick
  Nguyen, Zhifeng Chen, Anjuli Kannan, Ron~J Weiss, Kanishka Rao, Katya Gonina,
  et~al.,
\newblock ``State-of-the-art speech recognition with sequence-to-sequence
  models,''
\newblock in {\em International Conference on Acoustics, Speech and Signal
  Processing (ICASSP)}. IEEE, 2017.

\bibitem{miao2015eesen}
Yajie Miao, Mohammad Gowayyed, and Florian Metze,
\newblock ``Eesen: End-to-end speech recognition using deep rnn models and
  wfst-based decoding,''
\newblock in {\em Workshop on Automatic Speech Recognition and Understanding
  (ASRU)}. IEEE, 2015.

\bibitem{soltau2016neural}
Hagen Soltau, Hank Liao, and Hasim Sak,
\newblock ``Neural speech recognizer: Acoustic-to-word {LSTM} model for large
  vocabulary speech recognition,''
\newblock in {\em Interspeech}. ISCA, 2017.

\bibitem{audhkhasi2017direct}
Kartik Audhkhasi, Bhuvana Ramabhadran, George Saon, Michael Picheny, and David
  Nahamoo,
\newblock ``Direct acoustics-to-word models for english conversational speech
  recognition,''
\newblock in {\em Interspeech}. ISCA, 2017.

\bibitem{sennrich2015neural}
Rico Sennrich, Barry Haddow, and Alexandra Birch,
\newblock ``Neural machine translation of rare words with subword units,''
\newblock in {\em Annual Meeting of the Association for Computational
  Linguistics}. ACL, 2016.

\bibitem{caruana1997multitask}
Rich Caruana,
\newblock ``Multitask learning,''
\newblock {\em Machine Learning}, 1997.

\bibitem{audhkhasi2017building}
Kartik Audhkhasi, Brian Kingsbury, Bhuvana Ramabhadran, George Saon, and
  Michael Picheny,
\newblock ``Building competitive direct acoustics-to-word models for english
  conversational speech recognition,''
\newblock in {\em International Conference on Acoustics, Speech and Signal
  Processing (ICASSP)}. IEEE, 2018.

\bibitem{bahdanau2016end}
Dzmitry Bahdanau, Jan Chorowski, Dmitriy Serdyuk, Philemon Brakel, and Yoshua
  Bengio,
\newblock ``End-to-end attention-based large vocabulary speech recognition,''
\newblock in {\em International Conference on Acoustics, Speech and Signal
  Processing (ICASSP)}. IEEE, 2016.

\bibitem{graves2006connectionist}
Alex Graves, Santiago Fern{\'a}ndez, Faustino Gomez, and Jurgen Schmidhuber,
\newblock ``Connectionist temporal classification: labelling unsegmented
  sequence data with recurrent neural networks,''
\newblock in {\em International Conference on Machine learning}. ACM, 2006.

\bibitem{zenkel2017subword}
Thomas Zenkel, Ramon Sanabria, Florian Metze, and Alex Waibel,
\newblock ``Subword and crossword units for {CTC} acoustic models,''
\newblock in {\em International Conference on Acoustics, Speech and Signal
  Processing (ICASSP)}. ISCA, 2017.

\bibitem{chan2016latent}
Quoc V. Le Navdeep~Jaitly William~Chan, Yu~Zhang,
\newblock ``Latent sequence decompositions,''
\newblock {\em International Conference on Learning Representations (ICLR)},
  2016.

\bibitem{liu2017gram}
Hairong Liu, Zhenyao Zhu, Xiangang Li, and Sanjeev Satheesh,
\newblock ``Gram-{CTC}: Automatic unit selection and target decomposition for
  sequence labelling,''
\newblock in {\em International Conference on Machine Learning (ICML)}, 2017.

\bibitem{hori2017multi}
Takaaki Hori, Shinji Watanabe, and John~R Hershey,
\newblock ``Multi-level language modeling and decoding for open vocabulary
  end-to-end speech recognition,''
\newblock in {\em Workshop Automatic Speech Recognition and Understanding
  Workshop (ASRU)}. IEEE, 2017.

\bibitem{fiscus1997post}
Jonathan~G Fiscus,
\newblock ``A post-processing system to yield reduced word error rates:
  Recognizer output voting error reduction (rover),''
\newblock in {\em Workshop on Automatic Speech Recognition and Understanding
  (ASRU)}. IEEE, 1997.

\bibitem{kim2017joint}
Suyoun Kim, Takaaki Hori, and Shinji Watanabe,
\newblock ``Joint {CTC}-attention based end-to-end speech recognition using
  multi-task learning,''
\newblock in {\em International Conference on Acoustics, Speech and Signal
  Processing (ICASSP)}. IEEE, 2017.

\bibitem{dalmia2018sequence}
Siddharth Dalmia, Ramon Sanabria, Florian Metze, and Alan~W Black,
\newblock ``Sequence-based multi-lingual low resource speech recognition,''
\newblock in {\em International Conference on Acoustics, Speech and Signal
  Processing (ICASSP)}. IEEE, 2018.

\bibitem{sogaard2016deep}
Anders S{\o}gaard and Yoav Goldberg,
\newblock ``Deep multi-task learning with low level tasks supervised at lower
  layers,''
\newblock in {\em Annual Meeting of the Association for Computational
  Linguistics (ACL)}. ACL, 2016.

\bibitem{fernandez2007sequence}
Santiago Fern{\'a}ndez, Alex Graves, and J{\"u}rgen Schmidhuber,
\newblock ``Sequence labelling in structured domains with hierarchical
  recurrent neural networks,''
\newblock in {\em International Joint Conference on Artificial Intelligence
  (IJCAI)}, 2007.

\bibitem{toshniwal2017multitask}
Shubham Toshniwal, Hao Tang, Liang Lu, and Karen Livescu,
\newblock ``Multitask learning with low-level auxiliary tasks for
  encoder-decoder based speech recognition,''
\newblock in {\em Interspeech}. ISCA, 2017.

\bibitem{krishna2018hierarchical}
Kalpesh Krishna, Shubham Toshniwal, and Karen Livescu,
\newblock ``Hierarchical multitask learning for {CTC}-based speech
  recognition,''
\newblock {\em arXiv preprint arXiv:1807.06234}, 2018.

\bibitem{gage1994new}
Philip Gage,
\newblock ``A new algorithm for data compression,''
\newblock in {\em The C Users Journal}. R \& D Publications, Inc., 1994.

\bibitem{mikolov2012subword}
Tom{\'a}{\v{s}} Mikolov, Ilya Sutskever, Anoop Deoras, Hai-Son Le, Stefan
  Kombrink, and Jan Cernocky,
\newblock ``Subword language modeling with neural networks,''
\newblock {\em preprint (http://www. fit. vutbr. cz/imikolov/rnnlm/char. pdf)},
  2012.

\bibitem{zenkel2017comparison}
Thomas Zenkel, Ramon Sanabria, Florian Metze, Jan Niehues, Matthias Sperber,
  Sebastian St{\"u}ker, and Alex Waibel,
\newblock ``Comparison of decoding strategies for {CTC} acoustic models,''
\newblock in {\em Interspeech}. ISCA, 2017.

\bibitem{hochreiter1997long}
Sepp Hochreiter and J{\"u}rgen Schmidhuber,
\newblock ``Long short-term memory,''
\newblock {\em Neural Computation}, 1997.

\bibitem{hwang2016character}
Kyuyeon Hwang and Wonyong Sung,
\newblock ``Character-level incremental speech recognition with recurrent
  neural networks,''
\newblock in {\em International Conference on Acoustics, Speech and Signal
  Processing (ICASSP)}. IEEE, 2016.

\bibitem{povey2011kaldi}
Daniel Povey, Arnab Ghoshal, Gilles Boulianne, Lukas Burget, Ondrej Glembek,
  Nagendra Goel, Mirko Hannemann, Petr Motlicek, Yanmin Qian, Petr Schwarz,
  et~al.,
\newblock ``The kaldi speech recognition toolkit,''
\newblock in {\em Workshop on Automatic Speech Recognition and Understanding
  (ASRU)}. IEEE, 2011.

\bibitem{zeyer2018improved}
Albert Zeyer, Kazuki Irie, Ralf Schl{\"u}ter, and Hermann Ney,
\newblock ``Improved training of end-to-end attention models for speech
  recognition,''
\newblock in {\em Interspeech}. ISCA, 2018.

\end{thebibliography}

\end{document}